\documentclass[final]{article}

\usepackage[final]{neurips_2024}

\usepackage{graphicx}
\usepackage[utf8]{inputenc} % allow utf-8 input
\usepackage[T1]{fontenc}    % use 8-bit T1 fonts
\usepackage{hyperref}       % hyperlinks
\usepackage{url}            % simple URL typesetting
\usepackage{booktabs}       % professional-quality tables
\usepackage{amsfonts}       % blackboard math symbols
\usepackage{nicefrac}       % compact symbols for 1/2, etc.
\usepackage{microtype}      % microtypography
\usepackage{xcolor}         % colors

\title{Seeing Cells Clearly: Evaluating Machine Vision Strategies for Microglia Centroid Detection in 3D Images}

\author{Youjia Zhang \\
CSC 249: Machine Vision\\
University of Rochester\\
Rochester, NY 14627, USA\\
\texttt{yzh355@u.rochester.edu}
}

\begin{document}

\maketitle
\begin{abstract}
Microglia are important cells in the brain, and their shape can tell us a lot about brain health. In this project, I test three different tools for finding the center points of microglia in 3D microscope images. The tools include ilastik, 3D Morph, and Omnipose. I look at how well each one finds the cells and how their results compare. My findings show that each tool sees the cells in its own way, and this can affect the kind of information we get from the images.
\end{abstract}

\section{Introduction}
Microglia are immune cells that live in the brain. They change shape when the brain is hurt or under stress. By studying these changes, scientists can learn more about diseases like Alzheimer’s and autism. Today, we can take very detailed 3D pictures of microglia, but looking at all the images by hand takes too long. That is why we need good tools to help us find and study microglia automatically.

\section{Background (Related Work)}
Microglia are immune cells in the brain that change shape in response to disease, injury, and stress. These changes can show how the brain is reacting and are often used to study conditions like Alzheimer's disease, schizophrenia, and chronic stress (VonKaenel, 2023; Morrison et al., 2017). Even in healthy brains, microglia help remove dead cells and extra neural connections (Bilimoria and Stevens, 2015). To study these shapes, researchers use confocal microscopy to take high-resolution 3D images. These images are made of many 2D slices stacked together, but they often contain noise and many overlapping cells, which makes it hard to study one cell at a time.

\begin{figure}[h]
\centering
\includegraphics[width=0.6\textwidth]{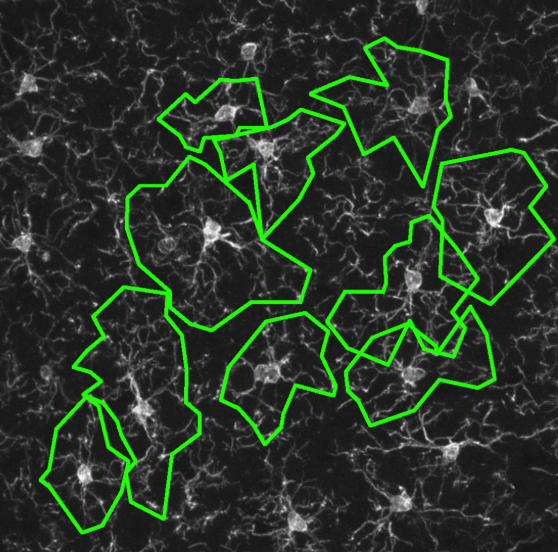}
\caption{Example of microglia segmentation on a single confocal image. Green contours outline detected cell boundaries. This illustrates how segmentation algorithms isolate individual microglia from dense, noisy data.}
\label{fig:microglia_segmentation}
\end{figure}

To deal with this, I used three tools that help find and segment microglia automatically. Ilastik uses machine learning to label pixels based on user input (Berg et al., 2019). 3D Morph uses basic image processing to find cell shapes and measure features like volume and branching (York et al., 2018). Omnipose uses deep learning to find the flow toward cell centers and group pixels into individual cells. Each tool gives different results and makes different assumptions. My goal was to compare these tools on the same dataset and see how well they find the center of each cell. A representative result showing segmented microglia is shown in Figure~\ref{fig:microglia_segmentation}.

\section{Methods}
\subsection{Dataset}
\begin{figure}[h]
\centering
\includegraphics[width=0.6\textwidth]{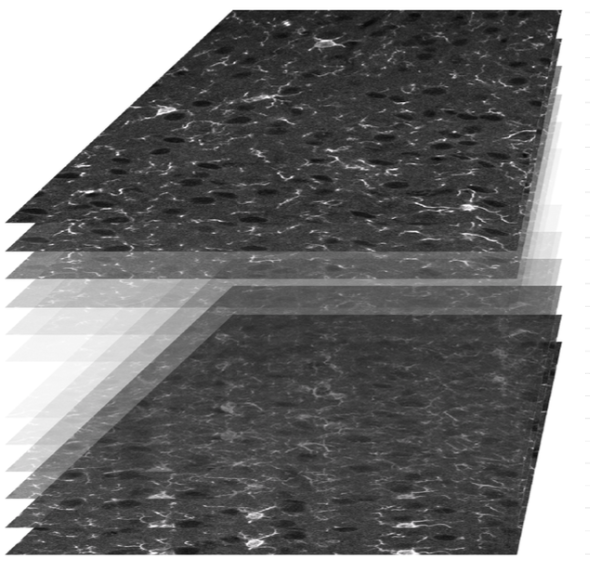}
\caption{Z-stack structure from the 12hrMD\_denoised dataset. Each z-stack is composed of 2D confocal slices taken at different depths to form a 3D image of in situ mouse microglia.}
\label{fig:zstack_dataset}
\end{figure}

The dataset I used is called \textbf{12hrMD\_denoised}. It comes from the McCall Research Group at the University of Rochester and contains 3D confocal microscopy images of in situ microglia in mouse brain tissue. These images were collected as z-stacks, where each z-stack is a series of 2D slices captured at different depths in the same biological volume (Figure~\ref{fig:zstack_dataset}).

In total, the dataset includes 1504 image slices, each with a resolution of 1024 by 1024 pixels. These slices are grouped into 103 z-stacks. Out of these, 1345 slices were used for training a denoising model, and 159 slices were used for testing. The dataset was processed using a self-supervised denoising algorithm that assumes a Poisson-Gaussian noise model. This model is well-suited for fluorescent microscopy data, where both signal-dependent and signal-independent noise are common.

The denoising model was trained over about 7 days using an 8-core CPU. The goal of this preprocessing step was to improve the signal-to-noise ratio in the raw microscopy images before running segmentation tools. For my experiments, I selected three representative z-stacks (numbers 71, 75, and 79) that cover different parts of the brain tissue and contain multiple visible microglia cells.

\subsection{Segmentation Tools}
\textbf{ilastik} is an interactive image segmentation tool that performs pixel classification using supervised machine learning. The main idea is to let users mark example pixels with simple brush strokes, labeling areas as either "cell" or "background." ilastik then calculates features like intensity, edge, and texture across different scales and uses a Random Forest classifier to predict which pixels belong to which class. The results appear in real time as a probability map, allowing quick feedback and refinement during training.

What makes ilastik accessible is that it works well with only a few training labels. It does not require coding or deep learning experience, so biologists can use it easily. The tool is fast and flexible, and its memory-efficient design allows it to process large datasets interactively, even on regular computers. Users can save the trained classifier and apply it to new images later in batch mode. This makes ilastik especially useful when working with many similar datasets.

However, ilastik performs semantic segmentation, not instance segmentation. That means it tells you whether a pixel belongs to a class but not which object it belongs to. If two cells are touching, ilastik might treat them as one object. To separate individual cells, you need to do post-processing like connected components or watershed algorithms. Also, because it only uses pre-defined features and not learned ones, it may not capture very subtle image patterns (Berg et al., 2019).

\textbf{3D Morph} is a classical image analysis pipeline built in MATLAB that was designed for analyzing microglial morphology in 3D. It starts by applying a threshold to the 3D image stack to separate bright regions from the background. These regions are then grouped into connected components, which are assumed to be individual cells. If a region looks too large, 3D Morph can split it into smaller parts using a Gaussian Mixture Model. This helps avoid merging nearby cells into one object.

After initial segmentation, 3D Morph creates a skeleton for each object. A skeleton is a thin structure that preserves the main shape and branching pattern of the cell. From this skeleton, the software calculates many useful features, including the number of endpoints, branch points, and average and maximum branch lengths. It also computes cell volume and territory volume, which helps describe how spread out the cell is in space.

The final results are saved in an easy-to-read table, which includes one row per cell with all its features. 3D Morph can be run in batch mode after you set the parameters, making it good for large datasets. It works best when the images are high quality and the cells are well separated. However, it may miss cells with faint signals or thin branches, especially if they do not survive the thresholding step. It also relies on setting the right parameters upfront, which can take some trial and error (York et al., 2018).

\textbf{Omnipose} is a deep learning model for segmentation that improves on earlier models like Cellpose. It works by predicting three maps from each image: a distance field showing how far each pixel is from the boundary, a boundary map, and a flow field that points from each pixel to the center of the cell. These maps are combined to decide which pixels belong together. Omnipose does not assume that cells are round or compact, so it works better on cells with odd shapes.

Omnipose uses a neural network with a U-Net structure, which is good at learning spatial features. It is trained on many different cell shapes, including long, curved, and branched cells. Once trained, it can generalize well to new images. The model separates touching cells more effectively than earlier methods because the flow field helps guide pixels toward the correct center. It also reduces errors in situations where other models often break cells into pieces.

In this project, I used a pre-trained Omnipose model for segmentation only. I did not retrain the model due to time limits. The omnipose gave good results on complex microglia shapes, but it does not directly provide structural features such as branch count or volume. These have to be calculated separately after segmentation. Training the model from scratch requires a lot of data and computing power. Still, for cases where shape complexity is high, Omnipose is often the most accurate option (Cutler et al., 2022).

\section{Experiment}

The goal of this project was to compare how well different image segmentation tools can detect the centroid of microglia cells in 3D confocal microscopy images. I focused on three tools: ilastik, 3D Morph, and Omnipose. Each tool uses a different strategy for segmentation, so I wanted to explore how their outputs differ when applied to the same dataset.

My main hypothesis was that each method would detect a different number of cells and that the results would vary in centroid accuracy, spatial distribution, and detail. I also expected that a deep learning model like Omnipose would perform better on complex cell shapes than traditional methods such as ilastik or 3D Morph.

To make the comparison fair, I used the same three z-stacks (numbered 71, 75, and 79) for all methods. These images were taken from a denoised 3D confocal microscopy dataset of mouse brain tissue. Each z-stack consists of a series of 2D image slices collected at different depths, forming a volumetric view of the sample.

For each method, I extracted three main types of output: the total number of detected cells, the centroid coordinates of each detected object, and, when available, additional shape-related features such as object size, volume, and branching statistics.

I evaluated each method using both quantitative and qualitative criteria. Quantitatively, I measured the average number of detected objects per z-stack. I also calculated the spatial spread of centroid coordinates to examine the degree to which the tools detected cells throughout the image. Qualitatively, I visualized the centroid locations to check for meaningful spatial patterns. I also reviewed whether the tools provided extra biological features, such as the ramification index or branch count. To assess accuracy, I compared the automated output with a manually traced reference set that included 81 human-labeled microglia paths.

This setup helped me explore how model design and segmentation strategy influence what each tool considers a cell and how well each tool can detect and represent the center of each one.

\section{Data Processing}
I selected three representative z-stacks from the data set: stacks 71, 75, and 79. For each stack, I processed the data using three different segmentation pipelines and generated one output file per method. These were:
\begin{itemize}
    \item \texttt{M\_} files: manually traced results
    \item \texttt{i\_} files: ilastik output
    \item \texttt{D\_} files: 3D Morph output
\end{itemize}

Each file stores feature-rich information for each detected object or path:
\begin{itemize}
    \item \texttt{M\_} files include path-level details such as path ID, name, coordinates (start and end in X, Y, and Z), path length, SWC type, primary and child path relationships, and fitted volume.
    \item \texttt{i\_} files contain predicted object-level data including object ID, predicted and user-assigned class, centroid coordinates (X and Y), bounding box ranges, pixel size, and class probabilities.
    \item \texttt{D\_} files offer rich morphological features including 3D centroid coordinates, object and territory volumes, ramification index, number of endpoints and branch points, and branch length statistics (average, maximum, and minimum).
\end{itemize}

These processed outputs allowed me to extract centroid information, count detected cells, and compare morphological features across methods for each z-stack.

\section{Results}
I applied ilastik, 3D Morph, and manual tracing to z-stack 79 and compared their outputs across several dimensions including the number of detected cells, centroid position, and morphological details.

\subsection{ilastik}
ilastik detected a total of 179 objects in the selected z-stack. The average 2D area of these segmented objects was approximately 1136 pixels. The segmentation results showed signs of over-segmentation, meaning that the algorithm may have broken larger structures into multiple smaller segments. This is consistent with ilastik’s design, which favors sensitivity over specificity to avoid missing weak signals.

\begin{figure}[h]
\centering
\includegraphics[width=0.6\textwidth]{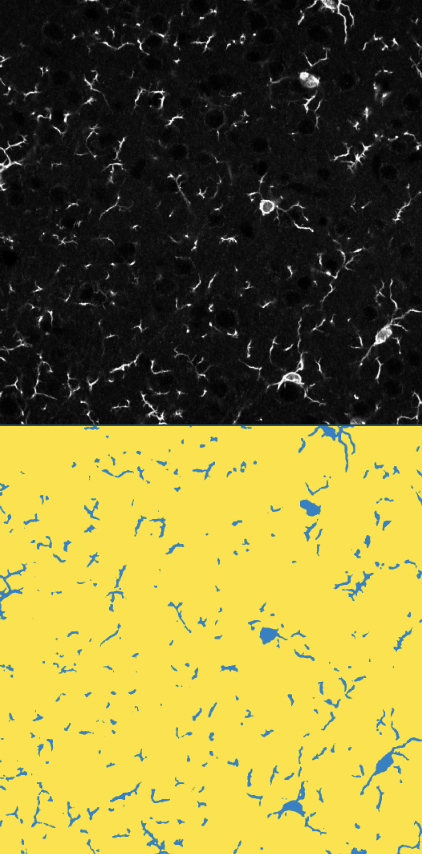}
\caption{ilastik segmentation result on z-stack 79. Top: original confocal image. Bottom: semantic segmentation output showing detected cells in blue against a yellow background. Some over-segmentation is visible in densely packed regions.}
\label{fig:ilastik_output}
\end{figure}

Although it performs well in identifying pixel classes, ilastik does not provide structural details such as branch count, volume, or ramification index. It produces semantic rather than instance segmentation, which limits the detail in morphological analysis. An example output showing semantic segmentation from ilastik is displayed in Figure~\ref{fig:ilastik_output}.

\subsection{3D Morph}
3D Morph detected only 11 well-defined 3D cell structures in the same dataset. The average volume of these objects was approximately 437~$\mu$m\textsuperscript{3}. Compared to ilastik, 3D Morph yielded fewer but cleaner results, likely due to its more conservative segmentation strategy that prioritizes connected voxel structures and removes weak or fragmented signals. This selective approach helps reduce false positives but may overlook fainter cells.

\begin{figure}[h]
\centering
\includegraphics[width=0.6\textwidth]{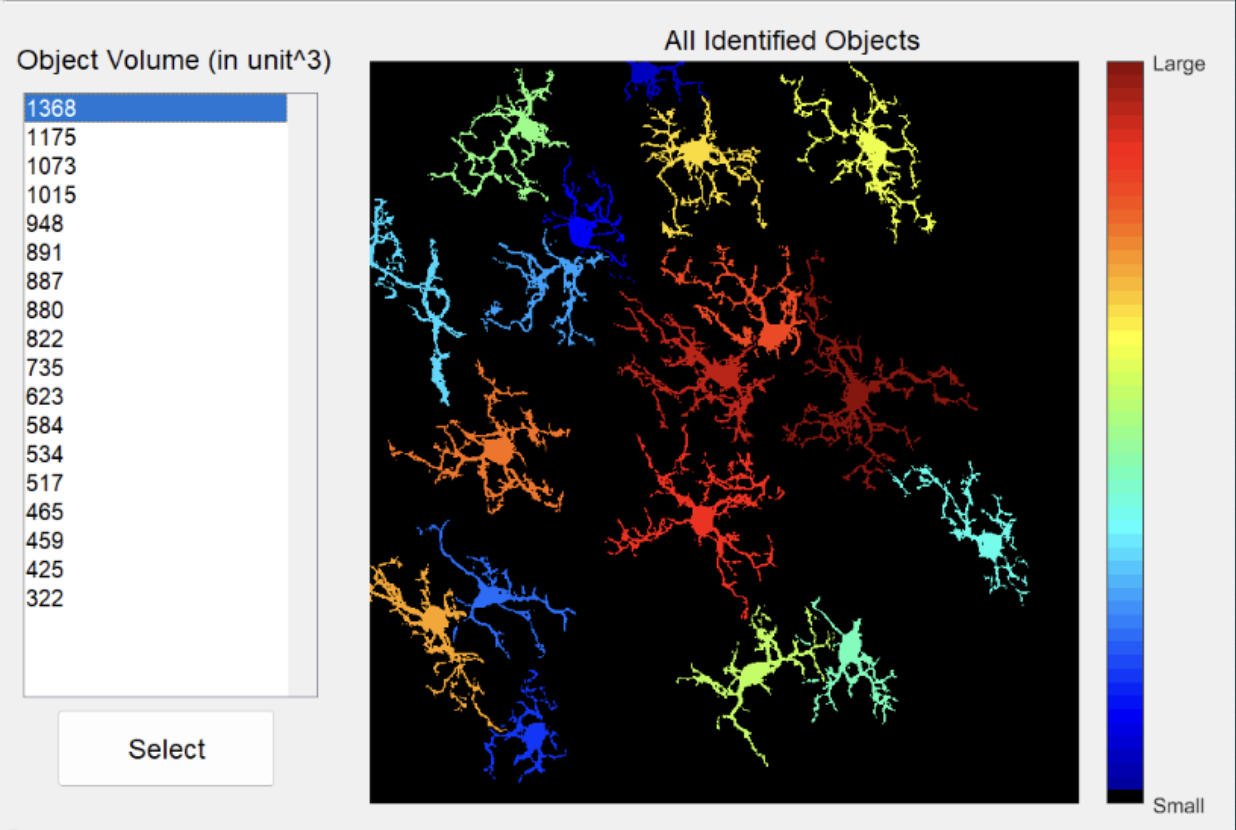}
\caption{3D Morph segmentation result on z-stack 79. Each identified microglia is color-coded by volume. Larger structures appear in red and orange, while smaller ones are shown in blue. The panel on the left lists volumes of detected objects in decreasing order.}
\label{fig:3dmorph_output}
\end{figure}

The output included a rich set of morphological features, such as the X, Y, and Z coordinates of each cell's centroid, object volume, ramification index, and the number of branch points and endpoints. This level of detail makes 3D Morph especially useful for morphology-based studies of microglia. The spatial distribution and relative sizes of the detected cells are visualized in Figure~\ref{fig:3dmorph_output}, where cells are color-coded based on their volumes.

To better understand the 3D spatial patterns of the detected cells, I extracted their centroid coordinates in microns. These values are listed in Table~\ref{tab:centroid_coordinates}, which shows variation across depth and lateral position within the z-stack. This data helps quantify how centralized or dispersed the detected objects are within the imaged region.
\begin{table}[h]
\centering
\caption{3D Centroid coordinates (in microns) from a selected z-stack}
\begin{tabular}{|r|r|r|}
\hline
\textbf{Centroid\_X\_µm} & \textbf{Centroid\_Y\_µm} & \textbf{Centroid\_Z\_µm} \\
\hline
13.96 & 60.70 & 10.30 \\
13.34 & 107.90 & 22.10 \\
31.61 & 12.73 & 20.29 \\
26.14 & 75.27 & 4.34 \\
33.37 & 130.36 & 5.51 \\
40.86 & 40.09 & 2.70 \\
67.34 & 88.69 & 18.98 \\
81.31 & 53.02 & 9.90 \\
71.32 & 60.54 & 23.08 \\
66.02 & 17.31 & 16.39 \\
77.67 & 118.37 & 29.73 \\
97.08 & 113.77 & 9.26 \\
102.06 & 17.08 & 12.60 \\
125.41 & 93.71 & 9.07 \\
98.75 & 64.56 & 28.30 \\
\hline
\end{tabular}
\label{tab:centroid_coordinates}
\end{table}

\subsection{Omnipose}
I also included manually traced data, which contained 81 paths drawn by a human observer. Volume measurements were not available because the optional volume field was left empty. However, manual tracing revealed more fine-scale structures and connectivity patterns that were not detected by either automated method. While this approach is highly accurate, it does not scale well and does not compute features like branch counts unless processed further.

\subsection{Centroid Position Comparison}
I compared centroid spread in the X and Y dimensions to assess spatial distribution patterns across segmentation methods. ilastik centroids were broadly distributed throughout the image, with X-coordinates ranging from 30.9 to 993.0 pixels and Y-coordinates from 14.3 to 985.9 pixels. The standard deviation was 274.2 pixels along the X-axis and 279.1 pixels along the Y-axis. This wide spread indicates that ilastik detects many small or loosely connected segments across the entire field of view, including objects near the image boundaries.

In contrast, 3D Morph centroids were more spatially concentrated. X-coordinates ranged from 7.6 to 163.9 microns, and Y-coordinates ranged from 19.4 to 152.1 microns. The standard deviation was 49.9 microns in X and 43.0 microns in Y. This tighter clustering reflects 3D Morph's emphasis on well-connected voxel structures and its tendency to avoid marginal or weak detections.

These results illustrate how different segmentation strategies lead to distinct spatial interpretations of the same biological data. The differences in centroid distribution, shown in Table~\ref{tab:centroid_comparison}, have direct consequences for downstream spatial and morphological analyses.
\begin{table}[h]
\centering
\caption{Centroid position comparison between ilastik and 3D Morph}
\begin{tabular}{|l|c|c|}
\hline
\textbf{Measurement} & \textbf{ilastik (pixels)} & \textbf{3D Morph (microns)} \\
\hline
X range & 30.9 – 993.0 px & 7.6 – 163.9 µm \\
Y range & 14.3 – 985.9 px & 19.4 – 152.1 µm \\
X standard deviation (spread) & 274.2 px & 49.9 µm \\
Y standard deviation (spread) & 279.1 px & 43.0 µm \\
\hline
\end{tabular}
\label{tab:centroid_comparison}
\end{table}
\section{Conclusion}
In this project, I compared three segmentation methods, namely ilastik, 3D Morph, and Omnipose, on the same 3D microglia dataset. Each method gave very different results based on how it was designed. ilastik found the most objects but often included noise. 3D Morph found fewer but cleaner objects and provided useful structural features. Omnipose handled complex shapes well, but I only used it for basic segmentation due to time limits.

These tools each have strengths and weaknesses. There is no one-size-fits-all solution, and the choice of method depends on what type of data you have and what you want to analyze. My comparison shows that segmentation tools can produce very different interpretations of the same image, so it is important to understand how each tool works before drawing biological conclusions.

\section{Implications}
The results show that different tools reflect different assumptions about what counts as a “cell.” ilastik uses a pixel-based classifier, so it tends to over-segment to avoid missing weak signals (Berg et al., 2019). 3D Morph extracts clean structures from thresholded 3D images, which gives better shape but fewer total detections (York et al., 2018). Manual tracing uses human judgment, which makes it better at catching fine-scale patterns that algorithms sometimes miss.

I also noticed that the spread of centroid positions depends on how each tool handles space. ilastik detects objects across the entire image, which can include false positives. 3D Morph limits detections to well-connected 3D shapes, reducing spatial noise. This suggests that model design does not just affect detection, but also influences how we understand biological structure. The choice of segmentation method can shape downstream analysis and even the scientific story that is told from the data.

\section{References}
{\small
\begin{itemize}
    \item Berg, S., Kutra, D., Kroeger, T. \textit{et al.}(2019). ilastik: interactive machine learning for (bio)image analysis. \textit{Nat Methods} 16, 1226–1232. https://doi.org/10.1038/s41592-019-0582-9
    \item Bilimoria, P. M., Stevens, B. (2015). Microglia function during brain development: New insights from animal models. *Brain Research*, 1617, 7–17.
    \item Cutler, K. J., Stringer, C., Lo, T. W., Rappez, L., Stroustrup, N., Brook Peterson, S., Wiggins, P. A., \& Mougous, J. D. (2022). Omnipose: A high-precision morphology-independent solution for bacterial cell segmentation. \textit{Nature Methods}, \textit{19}(11), 1438-1448. https://doi.org/10.1038/s41592-022-01639-4
    \item Morrison, H. W., Filosa, J. A., Tien, L. T. (2017). Microglia in health and disease. *Nature Reviews Neuroscience*, 18(3), 195–206.
    \item VonKaenel, E. D. (2023). *Methods for Microglia Image Analysis* (Doctoral dissertation, University of Rochester). Retrieved from University of Rochester Research Repository.
    \item York, N. R., Weigel, T. M., Svaren, J. (2018). 3DMorph: Automated analysis of microglia morphology in 3D. *Frontiers in Neuroinformatics*, 12, 101.

\end{itemize}
}

\end{document}